# Expensive Optimisation: A Metaheuristics Perspective

Maumita Bhattacharya
School of Computing & Mathematics
Charles Sturt University
NSW, Australia - 2640

*Abstract*—Stochastic, iterative search methods such as Evolutionary Algorithms (EAs) are proven to be efficient optimizers. However, they require evaluation of the candidate solutions which may be prohibitively expensive in many real world optimization problems. Use of approximate models or surrogates is being explored as a way to reduce the number of such evaluations. In this paper we investigated three such methods. The first method (DAFHEA) partially replaces an expensive function evaluation by its approximate model. The approximation is realized with support vector machine (SVM) regression models. The second method (DAFHEA II) is an enhancement on DAFHEA to accommodate for uncertain environments. The third one uses surrogate ranking with preference learning or ordinal regression. The fitness of the candidates is estimated by modeling their rank. The techniques' performances on some of the benchmark numerical optimization problems have been reported. The comparative benefits and shortcomings of both techniques have been identified.

## I. INTRODUCTION

Evolutionary Algorithms (EAs) are biologically inspired iterative processes where a population of candidate solutions is evolved generation after generation. In a typical EA a number of new offspring candidate solutions are produced through mutation, recombination and selection. Individuals for producing offspring are chosen using a selection strategy after evaluating the fitness value of each individual in the selection pool. In many real world optimization problems this fitness evaluation can be very expensive.

The use of surrogates to reduce the expensive function evaluation is found to be orders of magnitude cheaper computationally [21, 9, and 18]. Incorporation of approximate models may be one of the most promising approaches to realistically use EA to solve complex real life problems, especially where: (i). Fitness computation is highly time-consuming, (ii). Explicit model for fitness computation is absent, (iii). Environment of the evolutionary algorithm is noisy etc. However, considering the obvious risk involved in such approach, an EA with efficient control strategy for the approximate model and robust performance is welcome.

There are different ways, in which a surrogate or approximation model can be incorporated in an EA [15]; some of which are as follows:

*Problem level approximation*. In this approach, the statement of the problem itself is replaced by a reduced one that is easier to solve. See [15] for some examples on this.

*Functional approximation*. As the name suggests, in this approach, an alternate and explicit expression is constructed for the objective function, for the purpose of reducing the cost of evaluation. A set of evaluated points are used to build the approximate fitness model. This model is used to predict the fitness of candidate solutions. Usually a fraction of individuals in the population are selected and evaluated within each generation or over a number of generations to generate training points and are added to the training set to update the surrogates to maintain a reliable surrogate during evolution. See [13, 14, and 15] for examples on this technique.

*EA specific approximation*. This approach is specific for evolutionary algorithms and utilizes the algorithm's structural and functional aspects.

For a detailed review on use of approximation in EA, see [15]. In this paper we investigate three different methods which use surrogates to reduce the number of actual function evaluations in EA [4].

In the first one, namely, Dynamic Approximate Fitness based Hybrid Evolutionary Algorithm (DAFHEA), Bhattacharya et. al [2, 3] use both "functional approximation" and "EA specific approximation". It uses an approximation model to partially replace expensive fitness evaluations in evolutionary algorithm. DAFHEA uses an *explicit* control strategy (*a cluster-based on-line learning technique*) to improve reliability of using such approximate models to reduce expensive function evaluations. Also the approximate knowledge thus generated is exploited to avoid premature convergence (one of the major impediments of using evolutionary algorithm to solve complex real life optimization problems).

The second method, DAFHEA II [5] is an enhancement on DAFHEA to cover situations, where information from variable input dimensions and noisy data is involved. DAFHEA-II uses a multi-model regression approach. The multiple models are estimated by successive application of the SVM regression algorithm. Retraining of the model is done in a periodic fashion.

In the third method, Runersson [22] makes use of the EA feature that unlike classical optimization techniques, in rank based selection, selection of the best candidates requires only the rank or partial rank of the candidates. Here, the fitness of individuals is indirectly estimated by modeling their rank using surrogate. Preference learning or ordinal regression is used to implement a kernel-defined feature space.

The features and effectiveness of the above two surrogate-based methods have been investigated in this work. The above two methods have been selected for comparison as they are based on very different concepts and may reveal important characteristics which may be useful for specific problem cases.

Rest of the paper is organized as follows. Section II presents a brief review on use of surrogates in evolutionary computing. Section III outlines the features of the surrogate-based EA methods which we have investigated in this research. Section IV presents the experiment details and discussions on the findings. Finally, concluding remarks are summarized in Section V.

## II. Surrogate-Based Evolutionary Algorithm

The use of an approximate model to speed up optimization dates all the way back to the sixties [8]. The most widely used models being Response Surface Methodology [17], Krieging models [23] and artificial neural network models [6]. As has been mentioned in Section 1, the concept of using approximate model varies in levels of approximation (*Problem approximation, Functional approximation, and Evolutionary approximation*), model incorporation mechanism and model management techniques [15].

In the multidisciplinary optimization (MDO) community, primarily response surface analysis and polynomial fitting techniques are used to build the approximate models [11, 27]. These models work well when single point traditional gradient-based optimization methods are used. However, they are not well suited for high dimensional multimodal problems as they generally carry out approximation using simple quadratic models.

In another approach, multilevel search strategies are developed using special relationship between the approximate and the actual model. An interesting class of such models focuses on having many islands using low accuracy/cheap evaluation models with small number of finite elements that progressively propagate individuals to fewer islands using more accurate/expensive evaluations [29]. This approach may suffer from lower complexity/cheap islands having false optima whose fitness values are higher than those in the higher complexity/expensive islands. Rasheed et al. in [19, 20], uses a method of maintaining a large sample of points divided into clusters. Least square quadratic approximations are periodically formed of the entire sample as well as the big clusters. Problem of unevaluable points was taken into account as a design aspect. However, it is only logical to accept that true evaluation should be used along with approximation for reliable results in most practical situations. Another approach using population clustering is that of fitness imitation [15]. Here, the population is clustered into several groups and true evaluation is done only for the cluster representative [16]. The fitness value of other members of the same cluster is estimated by a distance measure. The method may be too simplistic to be reliable, where the population landscape is a complex, multimodal one.

Jin et al. in [13, 14] analyzed the convergence property of approximate fitness based evolutionary algorithm. It has been observed that incorrect convergence can occur due to false optima introduced by the approximate model. Two *controlled evolution* strategies have been introduced. In this approach, new solutions (offspring) can be (pre)-evaluated by the model. The (pre)-evaluation can be used to indicate promising solutions. It is not clear however, how to decide on the optimal fraction of the new individuals for which true evaluation should be done [1]. In an alternative approach, the optimum is first searched on the model. The obtained optimum is then evaluated on the objective function and added to the training data of the model [19, 26, and 1]. Yet another approach as proposed in [14], a regularization technique is used to eliminate false minima.

## III. The Investigated Methods

The main features of the three techniques investigated in this work, DAFHEA, DAFHEA II and the preference learning based EA are outlined below.

### A. The DAFHEA Technique

The primary objectives of the proposed algorithm and their realization are as below.

*1)* *The main objective of DAFHEA is to reduce the number of actual fitness function evaluations to speed up the search process. The proposed algorithm achieves this by partially replacing actual function evaluation (as is required in traditional genetic algorithm) by SVM based estimation. The DAFHEA framework includes a global model of genetic algorithm (GA), hybridized with support vector machine (SVM) [28] as the approximation tool.*

*2)* *The related major objective is to minimize the adverse effect of estimation. To this end explicit control strategies are used for evolution control, leading to considerable speedup without compromising heavily on solution accuracy.*

The *controlled* use of estimation is the primary reason why the proposed algorithm should be successful in reducing actual fitness function evaluation without heavily compromising on solution accuracy. The basic algorithm is as below.

**Step One:** Create a random population of $N_c$ individuals, where, $N_c = 5 * N_a$ and $N_a =$ actual initial population size.

**Step Two:** Evaluate $N_c$ individual using actual expensive function evaluation. Build the SVM approximate model using normalized expensive function evaluation values as training set for off-line training. (Use of normalized values in the training set appears to improve performance of meta-model, reducing effects of unnaturally high or low values). SVM hyper-parameters are initially tuned based on this training set.

**Step Three:** Select $N_a$ best individual out of $N_c$ evaluated individuals to form the initial GA population.

**Remarks:** The idea behind using five times the actual EA population size (as explained in Step One) is to make the approximation model sufficiently representative at least initially. Since initial EA population is formed with $N_a$ best individuals out of these $N_c$ individuals, with high recombination and low mutation rates, the EA population in first few generations is unlikely to drift much from its initial locality. Thus it is expected that large number of samples used in building the approximation model will facilitate better performance at this stage. Also using the higher fitness individuals, chosen out of a larger set should give an initial boost to the evolutionary process.

**Step Four:** Select parents using suitable selection operator and apply genetic operators namely recombination and mutation to create a new generation.

**Step Five:** Use SVM approximation model to compute fitness of new generation individuals based on approximate evaluation. Form $m$ distance-based (considering spatial distribution of individuals) clusters in the new population space. If for some $n$ clusters, the standard deviation $\sigma \geq$ Predefined Threshold, rearrange solution space into $m+n$ clusters. Compute a merit function $f_m(x)$ as below:

$$f_m(x) = f_a(x) - \rho_1 \sigma_i - \rho_2 d_{ij} \rho_3 s_i \quad (1)$$

In the equation (1), $f_a(x)$ is the predicted fitness function value. $\sigma_i$ is standard deviation (*in terms of objective value*) for the $i^{th}$ cluster and $d_{ij}$ is the normalized *minimum* Euclidean distance of $j^{th}$ point of $i^{th}$ cluster from the all truly evaluated points so far [22]. $s_i$ is the sparseness of the $i^{th}$ cluster. $\rho_1$, $\rho_2$ and $\rho_3$ are scaling factors for $\sigma_i$, $d_{ij}$ and $s_i$ respectively.

$$s_i = \frac{No \ of \ individuals \ in \ cluster \ i}{Dimension \ of \ individual} \quad (2)$$

**Step Six:** Dynamically update the approximate model as below:

1) Identify the cluster containing the optimum based on approximation.
2) Perform expensive evaluation for the approximate optimum and its $k-nearest$ neighbors.
3) Also perform expensive evaluation for the centroid of all other data clusters and their $k-nearest$ neighbors.
4) Expand neighborhood for true evaluation until a point is found in each space dimension such that percentage error $\delta \leq Predefined \ threshold$.

$$\delta = \left| \frac{a_{it} - a_{ip}}{a_{it}} \right| \times 100 \quad (3)$$

In the equation (3), $a_{it}$ =True value of the $i^{th}$ neighbor and $a_{ip}$ =Predicted value of the $i^{th}$ neighbor and **max** $i = k$.

Add the newly evaluated points to approximate model training set to update model.

**Step Seven:** When termination/evolution control criteria are not met, repeat Step Four to Step Seven.

**Remarks:** It must be noted, the optimum is considered based on the original predicted value $f_a(x)$. For all other purposes fitness based on the merit function $f_m(x)$ is considered. Periodic parameter tuning of the SVM approximation model was incorporated, though no specific criterion was used.

Further details on the above method can be found in [2, 3].

*B. The DAFHEA II Technique*

As in the original DAFHEA framework, DAFHEA-II [5] includes a global model of genetic algorithm (GA), hybridised with support vector machine (SVM) as the approximation tool. Expensive fitness evaluation of individuals as required in traditional evolutionary algorithm is partially replaced by SVM approximation models (unlike the original DAFHEA, multi-model regression is used). *Evolution control* is implemented by periodic true evaluations, leading to considerable speedup without compromising heavily on solution accuracy. Also the approximate knowledge about the solution space generated is used to maintain population diversity to avoid premature convergence.

*5) Functional Details*
The operational detail of DAFHEA-II [15] framework is as described below:

**Step One:** Create a random population of $N_c$ individuals, where, $N_c = 5 * N_a$ and $N_a$ = actual initial population size.

**Step Two:** Evaluate $N_c$ individual using actual expensive function evaluation. Build the SVM approximate models using the candidate solutions as input and the actual fitness (expensive function evaluation values) as targets forming the training set for *off-line training*.

**Step Three:** Select $N_a$ best individual out of $N_c$ evaluated individuals to form the initial GA population.

**Remarks:** The idea behind using five times the actual EA population size (as explained in *Step One*) is to make the approximation model sufficiently representative at least initially. Since initial EA population is formed with $N_a$ best individuals out of these $N_c$ individuals, with high recombination and low mutation rates, the EA population in first few generations is unlikely to drift much from its initial locality. Thus it is expected that large number of samples used in building the approximation model will facilitate better performance at this stage. Also using the higher fitness individuals, chosen out of a larger set should give an initial boost to the evolutionary process.

**Step Four:** Rank the candidate solutions based on their fitness value.

***Step Five:*** Preserve the elite by carrying over the best candidate solution to the next generation.

***Step Six:*** Select parents using suitable selection operator and apply genetic operators namely recombination and mutation to create children (new candidate solutions) for the next generation.

***Step Seven:*** The SVM regression models created in Step two are applied to estimate the fitness of the children (new candidate solutions) created in Step six. This involves assignment of most likely or appropriate models to each candidate solution.

***Step Eight:*** The set of newly created candidate solutions is ranked based on their approximate fitness values.

***Step Nine:*** The best performing newly created candidate solution and the elite selected in Step five are carried to the population of the next generation.

***Step Ten:*** New candidate solutions or children are created as described in Step six.

***Step Eleven:*** Repeat Step seven to Step ten until either of the following condition is reached:

1. The predetermined maximum number of generations has been reached; or
2. The periodic retraining of the SVM regression models is due.

***Step Twelve:*** If the periodic retraining of the SVM regression models is due, this will involve actual evaluation of the candidate solutions in the current population. Based on this training data new regression models are formed. The algorithm then proceeds to execute Step four to Step eleven.

***Remarks:*** The idea behind using periodic retraining of the SVM regression models is to ensure that the models continue to be representatives of the progressive search areas in the solution space.

*C. The Preference Learning Based EA*

The second method is directly based on preference learning or ordinal regression based technique proposed by Runersson in [22] with the variation that we have used a genetic algorithm implementation instead of CMA-ES. This method is based on the assumption that in a stochastic and direct search method such as EA, ordinal regression should be able to offer adequate surrogates as only full or even partial ranking of the individuals or search points is sufficient for the selection process. Accordingly, the surrogate approach is considered as a preference learning task, where a candidate point $x_i$ is preferred over $x_j$ if $x_i$ has a higher fitness than $x_j$. The training set for the surrogate model is thus composed of pairs of points $(x_i, x_j)_k$ and a corresponding label $r_k \in [1,-1]$, taking the value +1 or -1 depending on whether $x_i$ has a higher fitness than $x_j$ or vice versa.

The technique used for preferential learning or ordinal regression is kernel based. See [Runersson] for details on the method of ordinal regression using kernel defined features.

Model selection in surrogate ranking involves appropriately choosing a suitable kernel and its parameters as well as the regulation parameter $C$ which controls the balance between model complexities and training errors. Choice of a suitable kernel is problem specific.

As the search progresses, different regions of the search space are sampled and the original surrogate ranking model may be insufficiently accurate for new regions of the search space. It is therefore extremely important to update the surrogate during evolution. We have followed the surrogate update method suggested by Runersson in [22]. The strategy involves estimating the ranking of a population of points using the current surrogate and identifying the highest ranking point. The point is then evaluated using the true fitness function and its rank is calculated. Accuracy of the surrogate is evaluated by comparing the estimated rank with the true rank. The point evaluated with true fitness function is added to the training set.

IV. EXPERIMENTS

*A. Experiment Details for DAFHEA*

It may be noted that the target problem domain for our proposed algorithm involves time consuming actual fitness function evaluation. This property or characteristic of the fitness function is *external to the EA process*. Hence, to verify DAFHEA's effectiveness, it is sufficient to verify if DAFHEA can effectively reduce the number of actual function evaluations without compromising on accuracy for any set of standard test functions. Considering this, the performance of the proposed algorithm has been tested on five classical benchmark test functions: namely, Spherical, Ellipsoidal, Schwefel, Rosenbrock, and Rastrigin. Description of the test functions are as given in [3]. These benchmark functions in the test suit are scalable and are commonly used to assess the performance of optimization algorithms [30]. For Spherical and Rastrigin the global minimum is $f(x) = 0$ at $\{x_i\}^n = 0$. Rosenbrock has a global minimum of $f(x) = 0$ at $\{x_i\}^n = 1$.

All simulations were carried out using the following assumptions: The population size of $10n$ was used for all the simulations, where $n$ is the number of variables for the problem; for comparison purposes three sets of input dimensions are considered; namely, $n = 5$, 10 and 20. For all cases, tenfold validation was done with the number of generations being 1000; the SVM regression models [8] were trained with *five times* the real GA population size initially.

All the simulation processes were executed using a Pentium® 4, 2.4GHz CPU processor for both DAFHEA and the Preference Learning based EA.

*B. Experiment Details for DAFHEA II*

Both non-noisy and noisy versions of the chosen benchmark functions have been used to test DAFHEA II. The *noisy versions* of the functions have been obtained as follows.

$$f_{Noisy}(\vec{x}) = f(\vec{x}) + N(\mu, \sigma^2)$$

Here, $N(\mu, \sigma^2)$ = Standard Normal (or Gaussian) distribution with mean, $\mu = 0$ and variance, $\sigma^2 = 1$. The probability density function $f(x; \mu, \sigma^2)$ is defined as follows.

$$f(x; \mu, \sigma^2) = \frac{1}{\sigma\sqrt{2\pi}} \exp\left(-\frac{(x-\mu)^2}{2\sigma^2}\right)$$

All simulations were carried out using the following experiment setup: The population size of $10n$ was used for all the simulations, where $n$ is the number of variables for the problem; for comparison purposes three sets of input dimensions are considered; namely, $n = 5, 10$ and 20. For all three cases, tenfold validation was done with the number of iterations being 1000 for all non-noisy versions of the test problems; the SVM regression models were trained with *five times* the real EA (GA in this case) population size initially. However, in case of the noisy versions of the test functions much larger number of iterations has been used to obtain acceptable level of accuracy of results. All the simulation processes were executed using a Pentium® 4, 2.4GHz CPU processor.

### C. Experiment Details for Performance Learning Based EA

Following Runersson's [22] method a 2-norm soft margin support vector machine (SVM) has been used and the technique has been implemented using a classical genetic algorithm. As mentioned earlier, choice of appropriate kernel is an important factor in the performance learning based EA. Runersson [22] has tried ordinal regression with different kernels and concluded that 4th order polynomial kernel produces the best results for the Rosenbrock's function. For the sake of fair comparison we have used the same kernel for this test function. For the Spherical function, the 2nd order polynomial kernel performed best. Gaussian distribution with variance $0.1^2$ has been used for the Rastrigin's function.

Training points have been generated using a standard normal distribution centered about the origins (global minima) of the respective test functions. 1000 testing points were generated in the same manner. Using 60 randomly sampled training points the surrogate model has been estimated by ordinal regression. The regulation parameter $C$ has been chosen as 1.0E6.

As the search zooms in on a local minimum, the search will benefit from use of different kernel [22]. As suggested by Runersson in [22] a Gaussian distribution with variance $0.1^2$ was used in case of the Rosenbrock's and the Spherical functions in similar situations.

The surrogate has been validated and updated as explained in Section 3.2, every second generation.

### D. Results and Discussions

Performances of the three investigated methods on non-noisy versions of Spherical, Ellipsoidal, Schwefel, Rosenbrock, and Rastrigin functions with $n = 5$, 10 and 20 have been demonstrated in Table I. We have not reported any information on the number of actual function evaluations required for DAFHEA II in Table I as by design this technique employs additional function evaluations to achieve better performance in noisy environment. To give an idea about its efficacy in the noisy environment, Table II presents the comparative performances of the canonical Genetic Algorithm, DAFHEA and DAFHEA II in terms of number of actual function evaluations required when tested on the noisy versions of the test functions.

As can be observed from these results, Preference Learning based EA seems to have an advantage in terms of "number of actual function evaluations" over DAFHEA. However, its performance in terms of "mean fitness" is just not comparable to that of DAFHEA in all nine test cases. Both methods found the classical Spherical function easier to tackle as compared to the Rosenbrock's and the Rastrigin's functions. For both algorithms the mean function values for the spherical functions were better than their Rosenbrock counterparts. However, it may appear that based on the number of function evaluations, the spherical function was much harder for DAFHEA to solve than its Rosenbrock counterpart of the same dimension. It must be noted that increase in number of iterations and thus increase in the number of actual function evaluation showed no improvement in case of the Rosenbrock's function. In general, both models gained on performance with increase in training set size.

As can be anticipated, performances of both techniques deteriorated with increase in problem dimensions. However, this deterioration is much higher in case of the Preference Learning based EA, where the results are practically unusable except in case of Spherical function. Increase in the number of true function evaluations does not seem to improve the situation.

Other general observations are as below:

Both DAFHEA and Preference Learning based EA are applicable to situations where no explicit or computable fitness function is available. However, the concept of using preference learning based surrogate ranking may show more flexibility in such scenarios.

In the Preference Learning based EA, surrogate ranking has been realized using kernel based ordinal regression. That means the method is easily adaptable to any data types as long as a suitable kernel can be defined for the specific problem at hand. However, this is both an advantage and a disadvantage as this means, sufficient knowledge of the characteristics of the problem is required which may be difficult in real world scenarios.

The preference learning based EA benefits from selection of different kernel while the search zooms in on a local minimum. However, this switch may impose some additional computational as well as decisional overhead.

Surrogate ranking with RBF kernel tended to suffer from overfitting and get stuck in local minima. Second order polynomial performed better in case of higher order Rosenbrock's function.

The major drawback of the preference learning based surrogate ranking seems to be its inefficiency in handing higher dimensional problems, which is a common situation for most real world optimization problems.

TABLE I. PERFORMANCES OF THE DAFHEA TECHNIQUE (**M1**), THE DAFHEA II TECHNIQUE (**M2**) AND THE PREFERENCE LEARNING BASED EA (**M3**) AS IMPLEMENTED ON SPHERICAL, ELLIPSOIDAL, SCHWEFEL, ROSENBROCK, AND RASTRIGIN FUNCTIONS WITH $n = 5, 10$ AND $20$. PERFORMANCE MEASURES HAVE BEEN EXPRESSED AS THE "MEAN FITNESS" AND THE "NUMBER OF ACTUAL FUNCTION EVALUATIONS".

| Function | Mean Fitness (M1) | Mean Fitness (M2) | Mean Fitness (M3) | No of Actual Function Evaluations (M1) | No of Actual Function Evaluations (M3) |
|---|---|---|---|---|---|
| *Rosenbrock(5)* | 1.789E-41 | 1.998E-38 | 1.1103E-0.7 | 7015 | 1200 |
| *Rosenbrock(10)* | 1.991E-39 | 1.918E-26 | 1.0005 | 6990 | 4000 |
| *Rosenbrock(20)* | 2.313E-36 | 1.901E-19 | 2.1108 | 21170 | 17000 |
| *Spherical(5)* | 1.138E-60 | 1.138E-56 | 1.0102E-7 | 21210 | 375 |
| *Spherical(10)* | 1.152E-58 | 1.588E-43 | 1.0081E-5.5 | 77520 | 1200 |
| *Spherical(20)* | 1.58E-55 | 1.388E-35 | 1.0125E-5.5 | 110420 | 2750 |
| *Ellipsoidal(5)* | 3.220E-57 | 3.412E-51 | 1.0000E-6.1 | 18500 | 400 |
| *Ellipsoidal(10)* | 3.271E-55 | 2.523E-39 | 1.0100E-5.5 | 65700 | 1500 |
| *Ellipsoidal(20)* | 2.209E-52 | 1.323E-32 | 1.0511E-4.5 | 95510 | 2900 |
| *Schwefel(5)* | 1.198E-54 | 1.911E-48 | 1.0001E-0.8 | 11500 | 2700 |
| *Schwefel(10)* | 1.199E-51 | 2.971E-38 | 0.9000 | 15000 | 5000 |
| *Schwefel(20)* | 1.023E-48 | 1.989E-31 | 2.0002 | 25100 | 18000 |
| *Rastrigin(5)* | 3.285E-5 | 3.322E-1 | 1.1901E-0.8 | 4550 | 1700 |
| *Rastrigin(10)* | 3.089E-3 | 3.388E-1 | 0.9899 | 7175 | 5000 |
| *Rastrigin(20)* | 1.324E-1 | 10.032 | 3.0011 | 28010 | 15000 |

## V. CONCLUSIONS

Use of surrogates may be the most realistic answer to problems an iterative, stochastic search process like EA faces while dealing with situations, where, true fitness computation is highly expensive, or explicit model for fitness computation is absent, or environment of the evolutionary algorithm is noisy and so on. In this research, we have investigated three surrogate-based EA methods which aim at addressing some of these problems. While the first two methods, DAFHEA and DAFHEA II are based on "functional approximation" and "EA specific approximation" (see Section I), the second method uses surrogate ranking by ordinal regression or preference learning. Experiment results have shown, while Preference Learning based EA has some cost advantage in terms of number of true function evaluations, DAFHEA clearly should be the choice where accuracy (mean fitness value) is of paramount importance. DAFHEA II that uses multi-model regression for surrogate generation, shows some advantage over original DAFHEA and Canonical GA when applied to noisy functions, in terms of solution accuracy (results have not been shown in this article). However, this comes at the expense of some extra overhead in terms of number of actual function evaluations.

TABLE II. PERFORMANCES OF THE CANONOCAL GA (**M1**), THE DAFHEA TECHNIQUE (**M2**) AND THE DAFHEA II TECHNIQUE (**M2**) AS IMPLEMENTED ON **NOISY VERSIONS** OF SPHERICAL, ELLIPSOIDAL, SCHWEFEL, ROSENBROCK, AND RASTRIGIN FUNCTIONS WITH $n = 5, 10$ AND $20$. THE PERFORMANCE MEASURE HAS BEEN EXPRESSED AS THE "NUMBER OF ACTUAL FUNCTION EVALUATIONS".

| Function | No of Actual Function Evaluations (M1) | No of Actual Function Evaluations (M2) | No of Actual Function Evaluations (M3) |
|---|---|---|---|
| *Rosenbrock(5)* | 35,000 | 9500 | 9000 |
| *Rosenbrock(10)* | 100,000 | 71250 | 71000 |
| *Rosenbrock(20)* | 500,000 | 290,500 | 290,000 |
| *Spherical(5)* | 100,000 | 59000 | 58000 |
| *Spherical(10)* | 100,000 | 76000 | 75000 |
| *Spherical(20)* | 500,000 | 300,500 | 300,000 |
| *Ellipsoidal(5)* | 100,000 | 59000 | 58000 |
| *Ellipsoidal(10)* | 100,000 | 85000 | 84500 |
| *Ellipsoidal(20)* | 250,000 | 81550 | 81500 |
| *Schwefel(5)* | 100,000 | 69000 | 68000 |
| *Schwefel(10)* | 100,000 | 65000 | 64500 |
| *Schwefel(20)* | 300,000 | 200,050 | 200,000 |
| *Rastrigin(5)* | 100,000 | 5500 | 5100 |
| *Rastrigin(10)* | 100,000 | 20500 | 20000 |
| *Rastrigin(20)* | 500,000 | 410,500 | 410,000 |